\documentclass[10pt,twocolumn,letterpaper]{article}

\usepackage[pagenumbers]{cvpr} % To force page numbers, e.g. for an arXiv version

\usepackage{graphicx}
\usepackage{amsmath}
\usepackage{amssymb}
\usepackage{booktabs}
\usepackage{bm}

\usepackage{stfloats}
\usepackage{float}
\usepackage{multirow}

\usepackage{pifont}
\newcommand{\cmark}{\ding{51}}
\newcommand{\xmark}{\ding{55}}
\usepackage[pagebackref,breaklinks,colorlinks]{hyperref}

\usepackage[capitalize]{cleveref}
\crefname{section}{Sec.}{Secs.}
\Crefname{section}{Section}{Sections}
\Crefname{table}{Table}{Tables}
\crefname{table}{Tab.}{Tabs.}

\def\cvprPaperID{1075} % *** Enter the CVPR Paper ID here
\def\confName{CVPR}
\def\confYear{2023}

\begin{document}

\title{Joint Representation Learning for Text and 3D Point Cloud}

\author{%
  Rui Huang$^{1}$\thanks{Equal contribution.}\ \ \ \ \
  Xuran Pan$^{1*}$\ \ \ \ \
  Henry Zheng$^{1}$\ \ \ \ \
  Haojun Jiang$^{1}$\\ 
  Zhifeng Xie$^{2}$\ \ \ \ \
  Shiji Song$^{1}$\ \ \ \ \
  Gao Huang$^{1}$\thanks{Corresponding author.} \\
    $^{1}$ Department of Automation, BNRist, Tsinghua University\ \ \ \
    $^{2}$ Tsinghua University\\
  \texttt{\footnotesize \{hr20, pxr18, jh-zheng22, jhj20\}@mails.tsinghua.edu.cn} \\
  \texttt{\footnotesize \{xzhf, shijis, gaohuang\}@tsinghua.edu.cn}
}
\maketitle
\newcommand{\nameshort}[1]{Text4Point}

%%%%%%%%% ABSTRACT
\begin{abstract}
Recent advancements in vision-language pre-training (\textit{e.g.}, CLIP) have shown that vision models can benefit from language supervision. 
While many models using language modality have achieved great success on 2D vision tasks, the joint representation learning of 3D point cloud with text remains under-explored due to the difficulty of 3D-Text data pair acquisition and the irregularity of 3D data structure. 
In this paper, we propose a novel Text4Point framework to construct language-guided 3D point cloud models. The key idea is utilizing 2D images as a bridge to connect the point cloud and the language modalities. The proposed Text4Point follows the pre-training and fine-tuning paradigm. During the pre-training stage, we establish the correspondence of images and point clouds based on the readily available RGB-D data and use contrastive learning to align the image and point cloud representations. Together with the well-aligned image and text features achieved by CLIP, the point cloud features are implicitly aligned with the text embeddings. Further, we propose a Text Querying Module to integrate language information into 3D representation learning by querying text embeddings with point cloud features. For fine-tuning, the model learns task-specific 3D representations under informative language guidance from the label set without 2D images. Extensive experiments demonstrate that our model shows consistent improvement on various downstream tasks, such as point cloud semantic segmentation, instance segmentation, and object detection. The code will be available
\href{https://github.com/LeapLabTHU/Text4Point}{here}.

\end{abstract}

%%%%%%%%% BODY TEXT
\section{Introduction}
\label{sec:intro}

Recent progress in large-scale vision-language pre-training~\cite{jia2021scaling, radford2021learning, yuan2021florence} has shown that integrating the language modality into vision tasks exhibits promising performance. The high-quality visual representations learned by pre-trained multi-modal models can be effectively transferred to downstream tasks, such as image classification~\cite{zhou2021learning, yuan2021florence}, object detection~\cite{gu2021open, zhong2022regionclip}, and semantic segmentation~\cite{rao2021denseclip, zhou2022extract}.

\begin{figure}[t]
\centering
\includegraphics[scale=0.6]{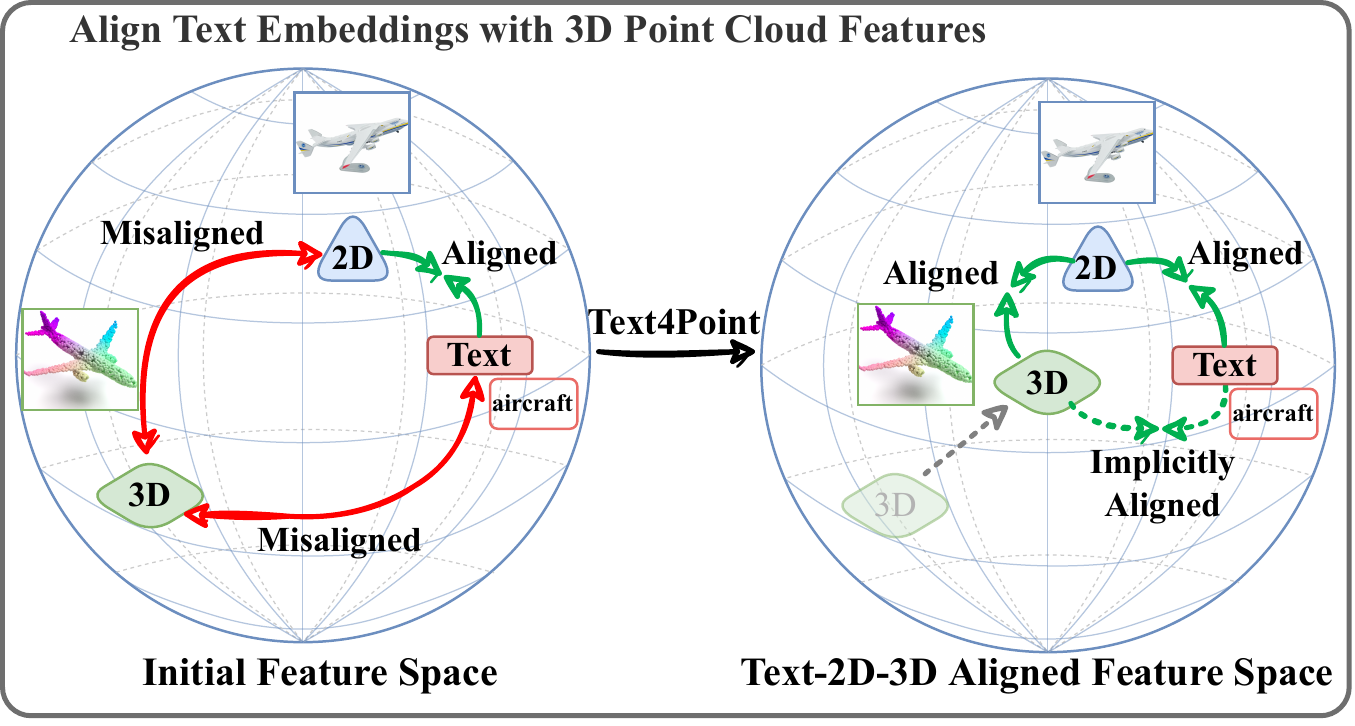}
\caption{Our proposed Text4Point uses 2D images as a bridge to align the embedding space of text and 3D point cloud data. With the help of CLIP~\cite{radford2021learning} which has aligned 2D features and text embeddings, we implicitly align 3D features and text embeddings by 2D-3D alignment.}
\label{fig:overview}
\vskip -0.3in
\centering
\end{figure}

The achievement in multi-modal joint learning has drawn much attention~\cite{gu2021open, rao2021denseclip, zhou2022towards, zhou2021learning, mokady2021clipcap, liu2022open, gao2021clip, patashnik2021styleclip} to introducing the language modality into 2D vision tasks. However, the area of combining text with 3D data is under-explored.
Recently, PointCLIP~\cite{zhang2021pointclip} has attempted to apply CLIP~\cite{radford2021learning} directly to 3D classification tasks~\cite{uy2019revisiting, wu20153d} by projecting point clouds to 2D depth images. Though effective, it has limitations in terms of applicability due to two reasons: 1) the multi-view projection process loses the 3D structural information, making the framework incompatible with fine-grained tasks like segmentation and detection; 2) the visual encoder of CLIP is not specifically designed for 3D tasks~\cite{dai2017scannet, wu20153d, song2015sun}, where depth map images projected from point clouds are usually out of the scope of the training data in CLIP, resulting in degraded performance. Therefore, the problem of designing general and flexible language-guided 3D models remains unsolved.

In fact, directly introducing the language modality into 3D vision models is a non-trivial task from two aspects. From the \textit{data} perspective, vision-language models put high demands on the scale of the pre-training dataset, \textit{e.g.}, 400 million image-text pairs for CLIP~\cite{radford2021learning}, which is difficult to meet under 3D scenarios. The cost of data acquisition in 3D vision is notably higher than capturing 2D images, not to mention the effort to generate text descriptions for point cloud data. 
Furthermore, the \textit{model structure and training paradigm} of 3D models~\cite{qi2017pointnet, qi2017pointnet++, wang2019dynamic} are dissimilar to that of language models due to the irregularity of the 3D data structure. The unique data pre-processing techniques, unit operators, and training objectives for 3D models make it difficult to be compatible with language models.

In this paper, we aim to bridge the gap between the point cloud and language modalities by leveraging 2D images as a link. In contrast to text, the alignment between 2D images and 3D point clouds is more straightforward and feasible. In fact, several previous works~\cite{qi2020imvotenet,chen2022autoalign} have already investigated improving the performance of 3D tasks with 2D prior knowledge, while other works~\cite{hou2021pri3d,chen2022self} successfully introduce 3D information into 2D tasks. With these advancements and given that various vision-language pre-training frameworks~\cite{radford2021learning, jia2021scaling} have already provided well-aligned 2D and language features, we propose a novel Text4Point method to align text embeddings with 3D point cloud features by using 2D data as the bridge.

Specifically, Text4Point follows the pre-training and fine-tuning paradigm. 
During pre-training, we utilize CLIP for the well-aligned language embeddings and 2D visual features. It is worth noting that 3D-Text data pairs are not required in our framework. Instead, we only use a list of label texts as the input of the language model. We align the encoded point cloud features to the CLIP-extracted image features by a contrastive loss. Hence, with the CLIP-learned Text-2D correspondence and our 2D-3D alignment, the text embeddings are implicitly aligned with the 3D features. Further, we introduce a Text Querying Module to extract language information by querying text embeddings with the aligned point cloud features. The extracted text features are then aggregated with the original geometric point cloud features to provide guidance for 3D representation learning.
For fine-tuning, with task-specific supervision, the model can directly learn 3D representations under language guidance from the label set without 2D images. 

Extensive experiments are conducted on various tasks to demonstrate the effectiveness of Text4Point. We perform semantic and instance segmentation on the widely used Stanford Large-Scale 3D Indoor Spaces (S3DIS)~\cite{armeni20163d} dataset and achieve consistent improvement over baselines.
For the object detection task, we evaluate our method on two popular benchmarks, SUN RGB-D\cite{song2015sun} and ScanNet\cite{dai2017scannet}.
Our framework exhibits +5.3\% and +4.3\% mAP@0.5 improvement for SUN RGB-D and ScanNet, respectively, and outperforms other competitive baselines.

\section{Related Works}
\noindent\textbf{Joint Representation Learning for Vision and Language.} 
There have been a series of works investigating joint representation learning for vision and language, \emph{e.g.}, image caption~\cite{xu2015show, anderson2018bottom, zhang2021vinvl}, text-to-image retrieval~\cite{wang2019camp, chen2020uniter, yuan2021florence}, visual question answering~\cite{antol2015vqa, su2019vl, yuan2021florence}, and so on. In the past few years, vision language pre-training has attracted more and more interest from researchers.
The most representative CLIP~\cite{radford2021learning} is trained with large-scale image-text data pairs via a contrastive objective and exhibits promising results in generalizing to various downstream tasks, \emph{e.g.}, image classification~\cite{radford2021learning, yuan2021florence}, semantic segmentation~\cite{rao2021denseclip, zhou2022extract}, image manipulation~\cite{patashnik2021styleclip, kwon2022clipstyler}, etc. 

Though great progress has been achieved in 2D vision-language pre-training, the area of joint representation learning of 3D vision and language is under-explored. 
Recently, PointCLIP~\cite{zhang2021pointclip} proposed to apply CLIP on 3D point cloud classification tasks. It projects the point cloud to 2D multi-view depth images, then directly deals with the 2D data. Since the original 3D structural information is discarded, it is limited to the classification tasks. Besides, the training data distribution of the CLIP image encoder is not consistent with the projected depth images, which may degrade the performance. Therefore, we put forward to connect language with 3D point cloud representations via 2D images.

\noindent\textbf{3D Dense Prediction Tasks.}
Similar to 2D dense prediction tasks, semantic segmentation~\cite{tchapmi2017segcloud, landrieu2018large, zhao2021point, guo2021pct, choy20194d}, instance segmentation~\cite{wang2018sgpn, jiang2020pointgroup, liang2021instance, wang2019associatively, yang2019learning}, and object detection~\cite{qi2020imvotenet, song2016deep, yang20203dssd, yin2021center, misra2021end} are also essential tasks in the 3D domain. 
Recent progress of 3D dense prediction tasks can be generalized into two types of approaches, point-based~\cite{qi2019deep, qi2020imvotenet, thomas2019kpconv, jiang2020pointgroup} and voxel-based~\cite{maturana2015voxnet, zhou2018voxelnet, yan2018second, choy20194d}. The point-based approaches mainly use PointNet~\cite{qi2017pointnet, qi2017pointnet++} or DGCNN~\cite{wang2019dynamic} networks to process point clouds. Voxel-based methods divide the point cloud into a 3D voxel grid, assign points to the corresponding voxel in the grid, and then use models such as sparse convolution networks~\cite{choy20194d} to process the voxels. 
The sparse convolution network, Sparse Residual UNet (SR-UNet)~\cite{choy20194d}, adopts UNet~\cite{ronneberger2015unet}-like architecture but converts the input data into a sparse tensor format and replaces the convolutional layers with sparse convolution tailored for sparse data. 
Due to the powerful feature extraction ability, SR-UNet achieves competitive performance on various 3D dense prediction tasks. Therefore, we adopt SR-UNet as the 3D backbone in our framework.
Since it is costly to annotate large-scale, high-quality data for dense prediction tasks, recent works~\cite{liu2021learning,yamada2022point,xie2020pointcontrast,rao2021randomrooms,wang2021unsupervised,yu2022point} explore to pre-train the 3D model for good representations which can be transferred to the downstream tasks. However, they focus on learning the geometrical 3D features while our framework introduces language information for 3D models to obtain better representations.

\section{Method}
In this section, we first review the model architecture design of CLIP in Sec.~\ref{sec:preliminaries} and give an overview of our Text4Point in Sec.~\ref{sec:overview}.
Then, we introduce our proposed pre-training framework in Sec.~\ref{sec:pretrain} and show how it can be applied to downstream tasks in Sec.~\ref{sec:finetune}.

\subsection{Preliminaries}
\label{sec:preliminaries}

Contrastive Language-Image Pre-training(CLIP)\cite{radford2021learning} framework learns visual concepts from natural language supervision. It optimizes an image encoder and a text encoder with a contrastive loss to predict the correspondence between images and the associated text descriptions. Concisely, CLIP extracts global features of image and text inputs, and computes the cosine distance as their similarity. Benefiting from the 400 million image-text pairs used in the training process, CLIP aligns the visual and language representations well. When transferring to downstream tasks, the text encoder of CLIP embeds the names of target classes under a pre-defined template, such as ``a photo of a [cls]". Then, CLIP predicts a probability distribution on all target categories between text embeddings and the image feature. By matching text embeddings with visual outputs, CLIP can perform reasonable predictions.
 
We first review the architecture of the ResNet-style CLIP image encoder that we adopt in our paper. It follows the ResNet~\cite{he2016deep} structure with an additional attention pooling layer following the last layer. Here, the feature of the last layer is denoted as $\bm{x} \in \mathbb{R}^{h\! \times\! w\! \times \!c}$, where $h$, $w$, and $c$ are height, width, and channel number of $\bm{x}$, respectively. 
The feature $\bm{x}$ is fed into a global average pooling layer and converted to a global feature $\bm{x_\textnormal{g}} \in \mathbb{R}^{1\! \times\! c}$. Then they are concatenated together and passed to a multi-head self-attention (MSA) layer. It can be formulated as:
\begin{align}
    \left[\bm{h}, \bm{h_\textnormal{g}}\right] = \textnormal{MSA}(\textnormal{Concat}\left(\bm{x}, \bm{x_\textnormal{g}}\right)),
\end{align}
where $\left[\bm{h}, \bm{h_\textnormal{g}}\right] \in \mathbb{R}^{(h\!\times\! w+1)\! \times\! d}$ is the output feature of the attention layer. The global token $\bm{h_\textnormal{g}}$ aggregates information from all the positions and is regarded as the representation of the whole image to match with text embeddings.

Transferring image features extracted by CLIP has been investigated in various vision tasks~\cite{zhou2021learning,rao2021denseclip,liu2022open,zhou2022towards,patashnik2021styleclip}. Most works follow the paradigm of CLIP and only use the global representations of images while neglecting the fine-grained features of the local regions.
However, for dense prediction tasks, fine features are essential~\cite{rao2021denseclip}. 
Therefore, we consider using the whole feature map $\bm{h}$ for 2D-3D pre-training.

\subsection{Overview of Text4Point}
\label{sec:overview}
\begin{figure}[htbp]
\centering
\includegraphics[scale=0.7]{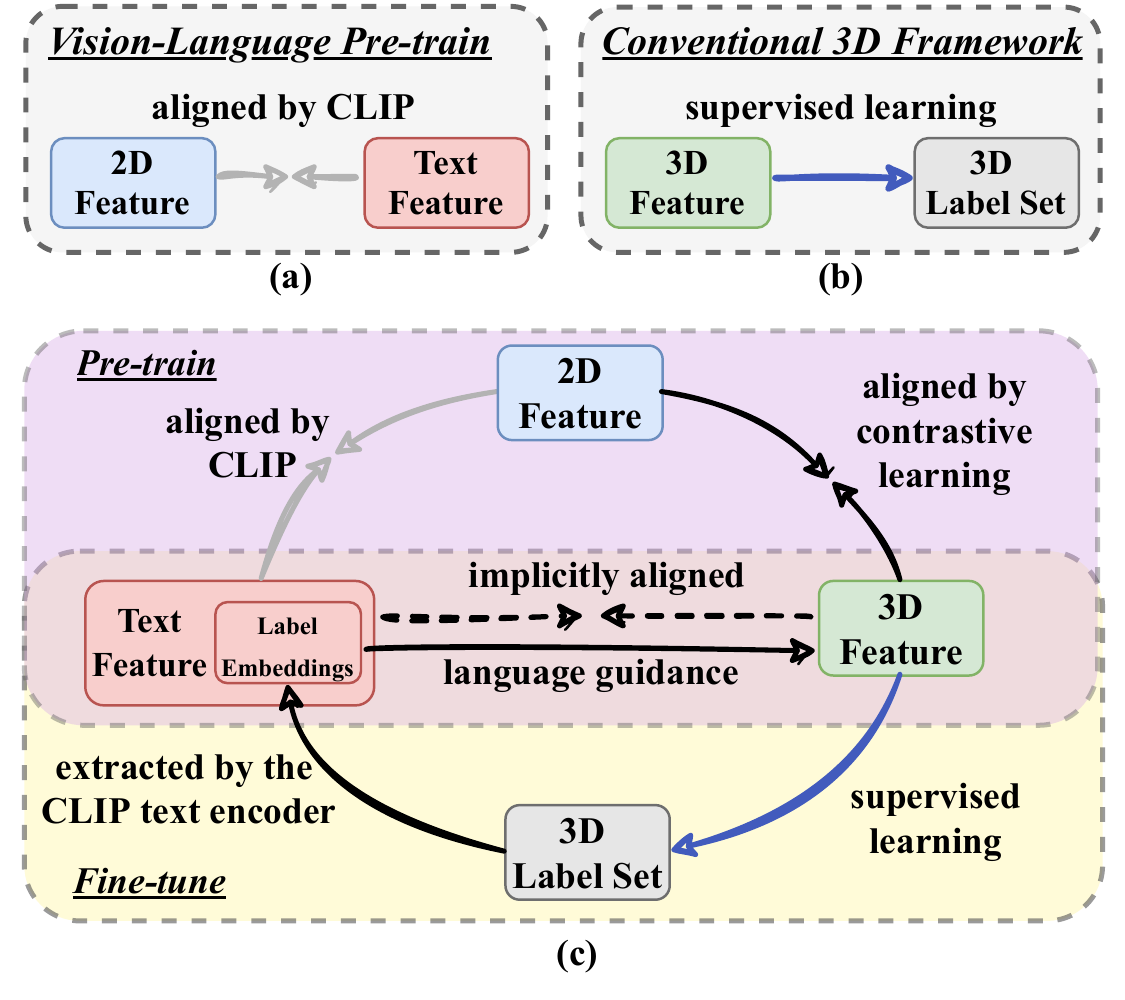}
\setlength{\abovecaptionskip}{0cm} 
\setlength{\belowcaptionskip}{0cm} 
\caption{Overview of our Text4Point. \textbf{(a)}: Vision-Language Pre-training aligns the feature space of text and 2D data. \textbf{(b)}: Conventional 3D framework only utilizes 3D information. \textbf{(c)}: Our framework connects the vision-language pre-training realm and the conventional 3D realm, which can provide language guidance for the point cloud representation learning.}
\label{fig:main_idea}
\vskip -0.2in
\centering
\end{figure}
To fully utilize language information in 3D representation learning, the domain gap between point cloud and text must be reduced. Since the pre-trained CLIP~\cite{radford2021learning} model has already aligned 2D visual features with text embeddings, it provides us an opportunity to bridge the gap between the features of point cloud and text. 

An overview of our Text4Point is shown in Fig.~\ref{fig:main_idea}(c). It follows the pre-training and fine-tuning paradigm. During pre-training , we establish connections between image and point cloud representations by aligning the embedding space of the CLIP image encoder and a 3D encoder. It is achieved through a contrastive pre-training process using 2D-3D data pairs obtained from the RGB-D dataset. With Text-2D and 2D-3D alignments, our method implicitly aligns the text representations and 3D features.
Meanwhile, we extract text embeddings for the classes in the pre-training dataset using the CLIP text encoder.
Furthermore, we introduce a Text Querying Module. It queries the text embeddings with the aligned point cloud features and can provide language guidance for the learning of the 3D decoder.
For fine-tuning, in addition to using the same supervision as the conventional 3D framework, we utilize the language guidance from the label embeddings generated by the label set for the 3D model. 

As illustrated in Fig.~\ref{fig:main_idea}, our framework connects the vision-language pre-training realm and the conventional 3D realm flexibly. New progress in these two realms can be integrated into our framework seamlessly, which would further improve the language-guided 3D model.

\subsection{Text4Point Pre-training}
\label{sec:pretrain}
\begin{figure*}[t]
    \centering
    \includegraphics[scale=0.6]{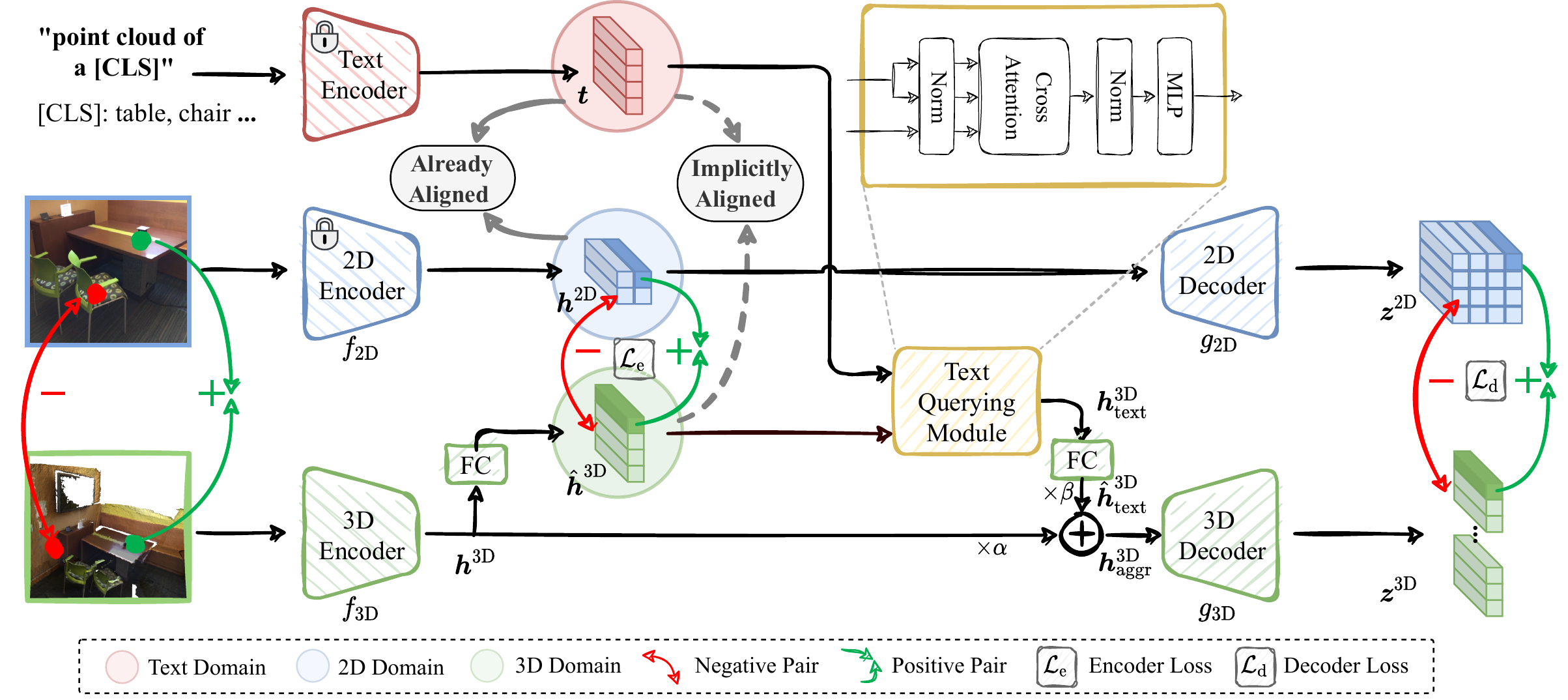}
    \caption{Illustration of the pre-training process. To enable the 3D model to leverage text information with the CLIP pre-trained text encoder, we align the embedding space of the CLIP pre-trained 2D encoder $f_{\textnormal{2D}}$ and our 3D encoder $f_{\textnormal{3D}}$ by a pixel-voxel contrastive loss.
    Further, a Text Querying Module is introduced to provide language guidance for the 3D model by querying text features with the aligned point cloud features. To obtain a well-initialized 3D model, we also align the feature of the 2D decoder $g_{\textnormal{2D}}$ and the 3D decoder $g_{\textnormal{3D}}$.}
    \label{fig:pretrain}
    \centering
    \vspace{-10pt}
\end{figure*}

\noindent\textbf{Data pre-processing.}
\label{sec:data_preprocess}
For pre-training, we build 2D-3D data pairs with the readily available RGB-D data. In specific, we use the ScanNet RGB-D dataset~\cite{dai2017scannet}, which contains sequences of color frames $\left\{ C_i\right\}$, depth frames $\left\{D_i\right\}$, and automatically-computed camera poses $\left\{T_i\right\}$. 
With the camera pose $T_i \in \mathbb{R}^{4\! \times\! 4}$ and camera intrinsic parameters $\mathcal{K} \in \mathbb{R}^{3\! \times\! 4}$, we can map the points from world coordinates to pixel coordinates and vice versa. Let ${\displaystyle [u\ v\ 1]^{\mathrm{T}}}$ represent a 2D point position in pixel coordinates, and ${\displaystyle [x_{\textnormal{w}}\ y_{\textnormal{w}}\ z_{\textnormal{w}}\ 1]^{\mathrm{T}}}$ represent its corresponding 3D point position in world coordinates. Formally, we have:
\begin{equation}
\label{eq:camera}
{\displaystyle [u\ v\ 1]^{\mathrm{T}}} = \mathcal{K} T_i {\displaystyle [x_{\textnormal{w}}\ y_{\textnormal{w}}\ z_{\textnormal{w}}\ 1]^{\mathrm{T}}}.
\end{equation}
Following \cite{hou2021pri3d}, at the $i$-th frame, we identify which point clouds in the world space correspond to it by back-projecting its depth map $D_i$ to camera space and mapping the computed points into world space with the camera pose $T_i$. The color frame $C_i$ and the corresponding point cloud $V_i$ form a 2D-3D data pair $(C_i,V_i)$.

\noindent\textbf{Model architecture.}
As shown in Fig.~\ref{fig:pretrain}, in the vision part of our architecture, we utilize 2D and 3D networks for pre-training, each having an encoder and a decoder. To align the embedding space of the CLIP pre-trained image encoder and the 3D encoder, we adopt the CLIP image encoder in our 2D network and combine it with a decoder consisting of bi-linear interpolation layers and convolutions. For the 3D network, we adopt the Sparse Residual UNet (SR-UNet)~\cite{choy20194d} architecture.
It has an encoder following the ResNet block structure and a decoder composed of convolutional layers. 
For pre-training, we randomly initialize the entire 3D model and the 2D decoder, and load the 2D encoder with the pre-trained CLIP weights.

We denote the encoders and decoders in 2D and 3D networks as $f_{\textnormal{2D}}$, $g_{\textnormal{2D}}$, $f_{\textnormal{3D}}$, $g_{\textnormal{3D}}$, and for each 2D-3D data pair $(C_i,V_i)$, the 2D image $C_i$ and its corresponding 3D point cloud $V_i$ are passed into the respective encoders. We have
\begin{equation}
    \label{eq:model}
    \bm{h}^{\textnormal{2D}}=f_{\textnormal{2D}}(C_i), \ \ \ 
    \bm{h}^{\textnormal{3D}}=f_{\textnormal{3D}}(V_i), 
\end{equation}
where $\bm{h}^{\textnormal{2D}} \in \mathbb{R}^{h\! \times\! w\! \times \!d}$ and $\bm{h}^{\textnormal{3D}} \in \mathbb{R}^{N\! \times\!d'}$ stand for the encoder feature of 2D and 3D.
Since the feature dimension of the CLIP image/text encoder $d$ is fixed, we expand the dimension of $\bm{h}^{\textnormal{3D}}$ to $d$ by a fully-connected (FC) layer to align the embedding space of the 2D encoder and 3D encoder:
\begin{equation}
    \label{eq:expand}
    \hat{\bm{h}}^{\textnormal{3D}}=\textnormal{FC}_{\textnormal{expand}}(\bm{h}^{\textnormal{3D}}),
\end{equation}
where $\hat{\bm{h}}^{\textnormal{3D}} \in \mathbb{R}^{N\! \times\!d}$ is the expanded 3D feature.

For the language part of our framework, we design a template as ``point cloud of [cls]." and construct $K$ text prompts for $K$ categories of the pre-training dataset. By feeding them into the CLIP pre-trained text encoder, we obtain the text embeddings $\bm{t} \in \mathbb{R}^{K\! \times \!d}$. 

\noindent\textbf{Text Querying Module.}
In order to obtain language guidance for 3D representation learning, we propose the Text Querying Module, which queries text embeddings with point cloud features and outputs a 3D-queried text feature.
Compared to the geometrical features extracted from the 3D encoder, language embeddings contain higher-level abstraction of visual concepts with rich semantic information. Thus, the 3D-queried feature can provide the model with valuable semantic guidance.

Specifically, the expanded 3D feature $\hat{\bm{h}}^{\textnormal{3D}}$ and the text embeddings $\bm{t}$ are normalized with batch normalization~\cite{ioffe2015batch} and performed cross attention, where $\hat{\bm{h}}^{\textnormal{3D}}$ and $\bm{t}$ serve as query and key/value, respectively. Then the obtained feature is fed into a multi-layer perceptron (MLP) to increase the nonlinearity. Mathematically, this can be formulized as:
\begin{gather}
    \label{eq:cross attention}
    \hat{\bm{h}}^{\textnormal{3D}}_{\textnormal{norm}}=\textnormal{Norm}(\hat{\bm{h}}^{\textnormal{3D}}), \ \ \  \bm{t}_\textnormal{norm}=\textnormal{Norm}(\bm{t}), \\
    \bm{Q}=\hat{\bm{h}}^{\textnormal{3D}}_{\textnormal{norm}}\bm{W}^{\bm{Q}},\bm{K}=\bm{t}_\textnormal{norm}\bm{W}^{\bm{K}}, \bm{V}=\bm{t}_\textnormal{norm}\bm{W}^{\bm{V}},\\
    \textnormal{Attn}(\bm{Q},\bm{K},\bm{V})=\textnormal{Softmax}\left(\bm{Q} \bm{K}^T/\sqrt{d}\right) \bm{V}, \\
    \bm{h}^{\textnormal{3D}}_{\textnormal{attn}}=\textnormal{Attn}(\bm{Q},\bm{K},\bm{V})\bm{W}^{\bm{O}},  \\
    \bm{h}^{\textnormal{3D}}_{\textnormal{text}}=\textnormal{MLP}(\textnormal{Norm}(\bm{h}^{\textnormal{3D}}_{\textnormal{attn}})),
\end{gather}
where $\bm{W}^{\bm{Q}}$,$\bm{W}^{\bm{K}}$,$\bm{W}^{\bm{V}}$,$\bm{W}^{\bm{O}}$ are parameter matrices, and $\bm{h}^{\textnormal{3D}}_{\textnormal{text}} \in \mathbb{R}^{N\! \times\!d}$ is the 3D-queried text feature.

To inject the language information provided by $\bm{h}^{\textnormal{3D}}_{\textnormal{text}}$ into the 3D model and facilitate the decoding process, we add the 3D-queried text feature to the original 3D encoder feature and send the aggregated feature to the 3D decoder. Before that, to prepare for the aggregation, we use a fully-connected layer to reduce the dimension of $\bm{h}^{\textnormal{3D}}_{\textnormal{text}}$ to the same as that of $\bm{h}^{\textnormal{3D}}$. Formally, this can be written as:
\begin{gather}
    \label{eq:aggregate}
    \hat{\bm{h}}^{\textnormal{3D}}_{\textnormal{text}}=\textnormal{FC}_{\textnormal{reduce}}(\bm{h}^{\textnormal{3D}}_{\textnormal{text}}), \\
    \bm{h}^{\textnormal{3D}}_{\textnormal{aggr}}=\alpha\bm{h}^{\textnormal{3D}}+\beta\hat{\bm{h}}^{\textnormal{3D}}_{\textnormal{text}}, \\
    \bm{z}^{\textnormal{3D}}=g_{\textnormal{3D}}(\bm{h}^{\textnormal{3D}}_{\textnormal{aggr}}),
\end{gather}
where $\hat{\bm{h}}^{\textnormal{3D}}_{\textnormal{text}} \in \mathbb{R}^{N\! \times\!d'}$, $\bm{h}^{\textnormal{3D}}_{\textnormal{aggr}} \in \mathbb{R}^{N\! \times\!d'}$ and $\bm{z}^{\textnormal{3D}} \in \mathbb{R}^{N'\! \times\!C}$ are the reduced 3D-queried feature, the aggregated 3D feature and the 3D decoder feature. The strength of aggregation is controlled by two learnable scalars $\alpha$ and $\beta$.
Meanwhile, we pass the 2D encoder feature $\bm{h}^{\textnormal{2D}}$ to the 2D decoder:
\begin{equation}
    \label{eq:decoder}
    \bm{z}^{\textnormal{2D}}=g_{\textnormal{2D}}(\bm{h}^{\textnormal{2D}}).
\end{equation}

\begin{figure*}[htbp]
\centering
\includegraphics[scale=0.6]{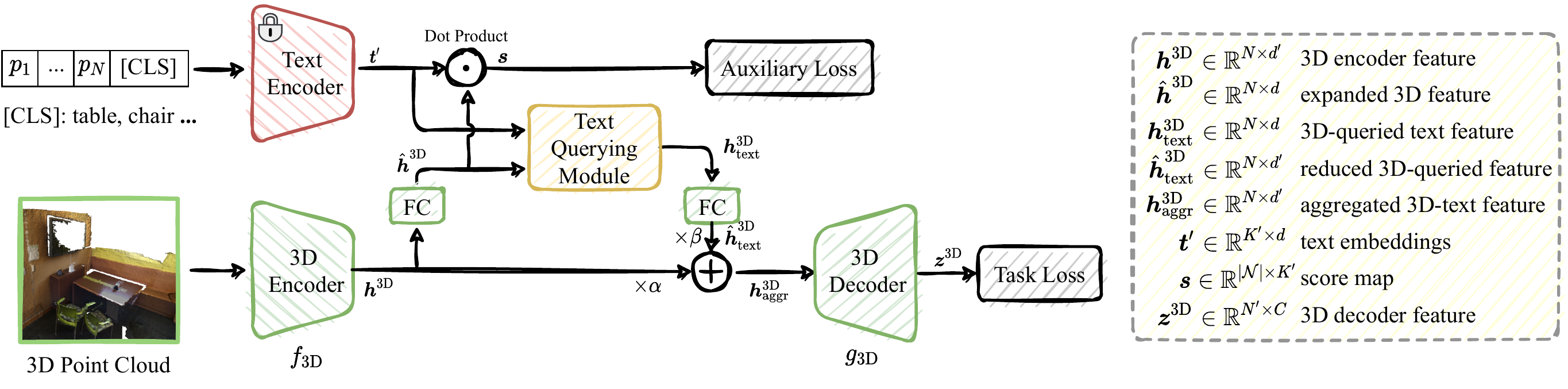}
\caption{Illustration of the fine-tuning process. We first extract $K'$-class text embeddings $\bm{t}'$ and 3D point cloud feature $\bm{h}^{\textnormal{3D}}$ with the CLIP text encoder and the 3D encoder $f_{\textnormal{3D}}$, respectively. Then we send them to the Text Querying Module and obtain the 3D-queried feature $\bm{h}^{\textnormal{3D}}_{\textnormal{text}}$ containing language information. It is aggregated with $\bm{h}^{\textnormal{3D}}$ to provide language guidance for the following decoding process.
The 3D network is fine-tuned with task-relevant supervision and an auxiliary loss.
}
\label{fig:finetune}
\centering
\vspace{-10pt}
\end{figure*}

\noindent\textbf{Training.}
Our goal is to enable the 3D model to leverage language information with the CLIP text encoder on the downstream tasks. Hence, we align the embedding space of the 3D encoder and the CLIP pre-trained image encoder. To preserve the good alignment between the CLIP encoders, we freeze their weights.

Inspired by CLIP, we adopt a contrastive learning method to implement the alignment between $\bm{h}^{\textnormal{2D}}$ and $\hat{\bm{h}}^{\textnormal{3D}}$. Since the local geometric features have proven to be critical for 3D point cloud tasks~\cite{xie2020pointcontrast}, we propose a pixel-voxel contrastive loss designed for representation learning.
First, we compute the pixel-voxel correspondences of $\bm{h}^{\textnormal{2D}}$ and $\hat{\bm{h}}^{\textnormal{3D}}$ according to Eq.~(\ref{eq:camera}). Coordinates for each voxel in $\hat{\bm{h}}^{\textnormal{3D}}$ are projected into camera space by the given camera pose and then transformed into the pixel coordinate system by the camera intrinsics. If a voxel lies within a certain distance of its corresponding pixel in the depth frame $D_i$ of the image $C_i$, they are considered a valid pixel-voxel pair. The distance is set to 2cm for pre-training the ScanNet.
We regard the pairs of correspondences as positive samples and all non-matching pixel-voxel pairs as negative samples. Formally, the contrastive loss is given by
\begin{equation}
\begin{split}
\label{loss}
\mathcal{L}_{\textnormal{e}}\!=&\!-\ \frac{1}{|S|} \sum_{(i, j) \in S}\!\!\! \left( \log \frac{\exp \! \left(\textnormal{cos}(\bm{h}^{\textnormal{2D}}_{i} \!\cdot \!\hat{\bm{h}}^{\textnormal{3D}}_{j}) / \tau\right)}{\sum_{(\cdot, p) \in S} \exp \! \left(\textnormal{cos}(\bm{h}^{\textnormal{2D}}_{i} \!\cdot \!\hat{\bm{h}}^{\textnormal{3D}}_{p})  / \tau\right)} \!  \right.\\ 
&\left. + \! \log \frac{\exp \! \left(\textnormal{cos}(\bm{h}^{\textnormal{2D}}_{i} \!\cdot \!\hat{\bm{h}}^{\textnormal{3D}}_{j}) / \tau\right)}{\sum_{(q, \cdot) \in S} \exp \! \left(\textnormal{cos}(\bm{h}^{\textnormal{2D}}_{q} \!\cdot \!\hat{\bm{h}}^{\textnormal{3D}}_{j})  / \tau\right)} \right),
\end{split}
\end{equation}
where $\textnormal{cos}(\cdot)$ denotes the cosine distance between the features, $\tau$ is the temperature hyper-parameter, and $S$ denotes the set of corresponding pixel-voxel pairs.

Meanwhile, to obtain a well-initialized 3D model, we also conduct a contrastive loss for the decoder feature of 2D and 3D. 
We can similarly calculate the contrastive loss $\mathcal{L}_{\textnormal{d}}$ of $\bm{z}^{\textnormal{2D}}$ and $\bm{z}^{\textnormal{3D}}$ by substituting $\bm{h}^{\textnormal{2D}}$ and $\hat{\bm{h}}^{\textnormal{3D}}$ in Eq.~(\ref{loss}) with $\bm{z}^{\textnormal{2D}}$ and $\bm{z}^{\textnormal{3D}}$. For simplicity, we do not repeat the calculation process here. 

\subsection{Text4Point Fine-tuning}
\label{sec:finetune}
\noindent\textbf{Language-guided prediction.}
As illustrated in Fig.~\ref{fig:finetune}, during the fine-tuning stage, we discard the 2D model, and only keep the 3D model and the CLIP pre-trained text encoder to perform language-guided prediction. They work in the same way as the pre-training process.
We first obtain the expanded 3D encoder feature $\hat{\bm{h}}^{\textnormal{3D}}$ and text embedding $\bm{t}' \in \mathbb{R}^{K'\! \times \!d}$, which is generated by the $K'$ class names of the downstream task. Then we feed them to the Text Querying Module and obtain the 3D-queried text feature $\bm{h}^{\textnormal{3D}}_{\textnormal{text}}$, which contains the language information. It is aggregated with the encoder feature $\bm{h}^{\textnormal{3D}}$ and passed to the 3D decoder. Finally, the decoder feature $\bm{z}^{\textnormal{3D}}$ is sent to the task-specific head for different downstream tasks.

\noindent\textbf{Prompt learning.}
By utilizing text prompts, CLIP closes the gap between the pre-training task and the downstream tasks. However, manually designed text templates are usually not optimal. 
Following~\cite{zhou2021learning}, we replace the fixed context in the text prompts with learnable parameters while freezing the entire weights of the text encoder. The model can optimize the continuous context vectors by back-propagation.
As the learnable prompts contain more precise descriptions of categories than handcrafted prompts, they can provide more useful language guidance and further improve our performance. It is worth noticing that once the training process is done, the text embeddings can be pre-calculated and used directly in the inference stage. In this way, our model only introduces the additional text encoder parameters in the training stage and keeps almost the same as the baseline model in the inference stage.

\noindent\textbf{Auxiliary loss.}
In fact, a data distribution shift exists between the pre-training and fine-tuning datasets. For instance, some categories in the downstream task may not have been seen in the pre-training stage. Since the text embeddings are generated according to the class names of the corresponding dataset, such disparity may hurt the transferability of the pre-trained model (especially the Text Querying Module, which utilizes the text embeddings). 
Therefore, we propose an auxiliary loss to mitigate this problem. 
With the text embedding $\bm{t}'$, we compute a rough classification score for each point in $\hat{\bm{h}}^{\textnormal{3D}}$:
\begin{equation}
    \label{eq:score_map}
    \bm{s}_{ij}\!=\!\textnormal{cos}(\hat{\bm{h}}^{\textnormal{3D}}_i,\bm{t}'^{\mathrm{T}}_j), i \!\in\!\{1, \ldots, n\},j \!\in\!\{1, \ldots, K'\},
\end{equation}
where $\bm{s} \in \mathbb{R}^{n\! \times\! K'}$ is a score map for 3D encoder feature.
Then we add an auxiliary supervision to it to provide more accurate features for the Text Querying Module. However, since the points in $\hat{\bm{h}}^{\textnormal{3D}}$ are not included in the input point set of $V_i$ due to sparse convolution~\cite{choy20194d}, the ground-truth labels of the points in $\hat{\bm{h}}^{\textnormal{3D}}$ cannot be referred to directly. Thus, we introduce a semantic label assignment module. Specifically, for each point in $\hat{\bm{h}}^{\textnormal{3D}}$, we check whether there exists any point, which we call ``neighbors," within a certain range around it. The auxiliary loss for the points in $\hat{\bm{h}}^{\textnormal{3D}}$ without ``neighbors" are skipped. For a point with ``neighbors," we use the statistical mode of the labels of its ``neighbors" as the pseudo label $y$ for it.

Let $\hat{\bm{s}}_i$ denote the score map scaled by a temperature $\tau$ and normalized via a softmax.
We adopt the standard cross-entropy loss function as the supervision of the score map:
\begin{equation}
    \label{eq:aux_loss}
    \mathcal{L}_{\textnormal{aux}}= \frac{1}{|\mathcal{N}|}\sum\nolimits_{i\in \mathcal{N}} L_{\textnormal{CE}}(\hat{\bm{s}}_i, {y}_i),
\end{equation}
where $\mathcal{N}$ denotes the set of the points in $\hat{\bm{h}}^{\textnormal{3D}}$ with ``neighbors" and $|\mathcal{N}|$ denotes the total number of points in $\mathcal{N}$.

\section{Experiment}
\label{sec:exp}
In this section, extensive experiments are conducted to validate our method on dense prediction tasks in 3D domain. 
We show that with the Text4Point framework, language information helps improve the performance of 3D models on various applications, including semantic segmentation, instance segmentation, and object detection. 
Ablation studies further demonstrate the necessity of introducing the language modality and the effectiveness of each component in our framework. 

\subsection{Pre-training}
\noindent\textbf{Setup.}
We utilize the ScanNet dataset \cite{dai2017scannet} for Text4Point pre-training, which is collected by a lightweight RGB-D scanning procedure. ScanNet is a collection of 1513 indoor scenes containing 2.5 million views, annotated with 3D camera poses, surface reconstructions, and instance-level semantic segmentations. 
We create a 2D-3D pair dataset on top of ScanNet for the pre-training framework as mentioned in Sec.~\ref{sec:data_preprocess}. Specifically, for each scene, we sub-sample every 25 frames of the RGB-D scan, and for each frame, we extract a 2D-3D data pair.

\noindent\textbf{Implementation details.}
We apply the ResNet-style CLIP as our 2D encoder and text encoder, and freeze its weights during training. The 2D decoder is composed of bi-linear interpolation layers and convolutional layers. We adopt the 34-layer Sparse Residual UNet as our 3D backbone. The 2D decoder and the 3D model are randomly initialized. To provide meaningful 2D features for the 3D decoder to align, we use the free depth images in the RGB-D dataset to supervise the learning of the 2D decoder. Thus, we adopt an auxiliary depth loss by default. Please refer to the Appendix for details. 
The input image is resized to $240\times 320$, and the voxel size is set as 2.5cm.
The temperature $\tau$ of pixel-voxel contrastive loss is set to 0.4. 
We use momentum SGD with an initial learning rate of 0.1 and choose the exponential scheduler with a power factor of 0.99. The model is trained for 2 epochs with a batch size of 32. 

\begin{table*}[t]\small
\setlength{\abovecaptionskip}{0cm} 
\setlength{\belowcaptionskip}{0cm} 
\caption{
\textbf{Semantic segmentation results on S3DIS.} 
We report the mean IoU and mean accuracy of all 13 classes, as well as the IoU for each object category. * corresponds to the model adopts SR-UNet~\cite{choy20194d} as the backbone.}
% \vskip -0.1in
\newcommand{\tabincell}[2]{\begin{tabular}{@{}#1@{}}#2\end{tabular}}
\begin{center}
\setlength{\tabcolsep}{0.6mm}{
\renewcommand\arraystretch{1.1}
\begin{tabular}{l|cc|ccccccccccccc}
\toprule
\multicolumn{16}{c}{\textbf{Semantic Segmentation on S3DIS}} \\
Method & mIoU & mAcc & ceiling & floor & wall & beam & column & window & door & chair & table & bookcase & sofa & board & clutter\\
\hline
PointNet \cite{qi2017pointnet} & 41.1 & 49.0 & 88.8 & 97.3 & 69.8 & 0.1 & 3.9 & 46.3 & 10.8 & 58.9 & 52.6 & 5.9 & 40.3 & 26.4 & 33.2\\
SegCloud \cite{tchapmi2017segcloud} & 48.9 & 57.4 & 90.1 & 96.1 & 69.9 & 0.0 & 18.4 & 38.4 & 23.1 & 75.9 & 70.4 & 58.4 & 40.9 & 13.0 & 41.6\\
TangentConv \cite{tatarchenko2018tangent} & 52.8 & 60.7 & 90.5 & 97.7 & 74.0 & 0.0 & 20.7 & 39.0 & 31.3 & 77.5 & 69.4 & 57.3 & 38.5 & 48.8 & 39.8\\
3D RNN \cite{ye20183d} & 53.4 & 71.3 & 95.2 & 98.6 & 77.4 & 0.8 & 9.8 & 52.7 & 27.9 & 76.8 & 78.3 & 58.6 & 27.4 & 39.1 & 51.0\\
% DGCNN \cite{wang2019dynamic} & 56.1 & - & - & - & - & - & - & - & - & - & - & - & - & - & -\\
PointCNN \cite{li2018pointcnn} & 57.3 & 63.9 & 92.3 & 98.2 & 79.4 & 0.0 & 17.6 & 22.8 & 62.1 & 74.4 & 80.6 & 31.7 & 66.7 & 62.1 & 56.7\\
SPGraph \cite{landrieu2018large} & 58.0 & 66.5 & 89.4 & 96.9 & 78.1 & 0.0 & 42.8 & 48.9 & 61.6 & 84.7 & 75.4 & 69.8 & 52.6 & 2.1 & 52.2 \\
PCNN \cite{atzmon2018point} & 58.3 & 67.0 & 92.3 & 96.2 & 75.9 & 0.3 & 6.0 & 69.5 & 63.5 & 66.9 & 65.6 & 47.3 & 68.9 & 59.1 & 46.2 \\
KPConv \cite{thomas2019kpconv} & 67.1 & 72.8 & 92.8 & 97.3 & 82.4 & 0.0 & 23.9 & 58.0 & 69.0 & 81.5 & 91.0 & 75.4 & 75.3 & 66.7 & 58.9 \\
PCT \cite{guo2021pct} &  61.3 & 67.7 & 92.5 & 98.4 & 80.6 & 0.0 & 19.4 & 61.6 & 48.0 & 76.6 & 85.2 & 46.2 & 67.7 & 67.9 & 52.3\\
Point Transformer \cite{zhao2021point} & 70.4 & 76.5 & 94.0 & 98.5 & 86.3 & 0.0 & 38.0 & 63.4 & 74.3 & 89.1 & 82.4 & 74.3 & 80.2 & 76.0 & 59.3 \\
\hline
SR-UNet \cite{choy20194d} & 68.2 & 75.5 & 91.5 & 98.6 & 84.1 & 0.0 & 33.0 & 56.9 & 64.0 & 90.1 & 81.7 & 72.5 & 76.5 & 77.9 & 59.6 \\
PointContrast \cite{xie2020pointcontrast}* & 70.3 (\textcolor{blue}{\scriptsize{+2.1}}) & 76.9 & 93.3 & 98.7 & 85.6 & 0.1 & 45.9 & 54.4 & 67.9 & 91.6 & 80.1 & 74.7 & 78.2 & 81.5 & 62.3 \\
\textbf{Text4Point}* & \textbf{72.2 (\textcolor{blue}{\scriptsize{+4.0}})} & \textbf{78.5} & 95.6 & 98.6 & 87.7 & 0.0 & 36.4 & 61.4 & 80.0 & 91.0 & 81.7 & 75.8 & 85.0 & 82.0 & 63.8 \\
\bottomrule
\end{tabular}}
\end{center}
\label{tab:semseg}
\vskip -0.2in
\end{table*}

\begin{table}[htbp]\small
\setlength{\abovecaptionskip}{0cm} 
\setlength{\belowcaptionskip}{0cm} 
\caption{\textbf{Instance segmentation results on S3DIS.} We report mean average precision (mAP) as well as mean precision (mPrec) and mean recall (mRec) with 0.5 IoU threshold. * corresponds to the model adopts SR-UNet~\cite{choy20194d} as the backbone.}
\newcommand{\tabincell}[2]{\begin{tabular}{@{}#1@{}}#2\end{tabular}}
\begin{center}
\setlength{\tabcolsep}{0.7mm}{
\renewcommand\arraystretch{1.1}
\begin{tabular}{l|cccc}
\toprule
\multicolumn{5}{c}{\textbf{Instance Segmentation on S3DIS}} \\
Method & mAP & mAP@0.5 & mPrec & mRec \\
\hline
SGPN~\cite{wang2018sgpn} & - & 54.4 & 36.0 & 28.7 \\
ASIS~\cite{wang2019associatively} & - & - & 55.3 & 42.4 \\
3D-Bonet~\cite{yang2019learning} & - & - & 57.5 & 40.2 \\
PointGroup~\cite{jiang2020pointgroup} & - & 57.8 & 61.9 & 62.1 \\
SSTNet~\cite{liang2021instance} & 42.7 & 59.3 & 65.5 & 64.2 \\
\hline
SR-UNet \cite{choy20194d} & - & 59.3 & - & - \\
PointContrast \cite{xie2020pointcontrast}* & - & 60.9 (\textcolor{blue}{\scriptsize{+1.6}}) & - & - \\
\textbf{Text4Point}* & \textbf{47.3} & \textbf{62.7 (\textcolor{blue}{\scriptsize{+3.4}})} & 63.9 & \textbf{68.2} \\
\bottomrule
\end{tabular}}
\end{center}
\label{tab:insseg}
\vskip -0.3in
\end{table}

\subsection{Semantic and Instance Segmentation}
\noindent\textbf{Setup.}
We perform segmentation tasks on Stanford Large-Scale 3D Indoor Spaces (S3DIS)~\cite{armeni20163d} dataset. It contains 3D scans of 6 large-scale indoor areas with 271 rooms. The point clouds of each scan are annotated with instance labels and semantic labels of 13 object categories. We fine-tune our pre-trained model following the training settings of PointContrast\cite{xie2020pointcontrast}. Please refer to the Appendix for details.
The commonly adopted benchmark split (“Area 5 test”) is adopted for training and testing.

\noindent\textbf{Results.}
The semantic and instance segmentation results on the S3DIS dataset are summarized in Tab.~\ref{tab:semseg} and Tab.~\ref{tab:insseg}, respectively. 
For semantic segmentation, we report the mean IoU and mean accuracy of all 13 classes. 
Text4Point achieves 72.2\% mIoU, surpassing PointContrast by 1.9\%. Besides, Text4Point significantly improves the vanilla SR-UNet by 4.0\% on mIoU and outperforms it over 10 out of 13 categories. 
For instance segmentation, we report mean average precision (mAP) as well as mean precision (mPrec) and mean recall (mRec) with 0.5 IoU threshold. By adopting Text4Point, SR-UNet obtains 3.4\% improvement on mAP@0.5 and outperforms other competitive baselines.

\begin{table}[htbp]\small
\setlength{\abovecaptionskip}{0cm} 
\setlength{\belowcaptionskip}{0cm} 
\caption{\textbf{Object detection results on SUN RGB-D.} We report the mean average precision with 0.5 IoU threshold. * corresponds to the model adopts SR-UNet~\cite{choy20194d} as the backbone. $\dag$ corresponds to the model adopts PointNet++~\cite{qi2017pointnet++} as the backbone.}
\newcommand{\tabincell}[2]{\begin{tabular}{@{}#1@{}}#2\end{tabular}}
\begin{center}
\setlength{\tabcolsep}{2.5mm}{
\renewcommand\arraystretch{1.1}
\begin{tabular}{l|c|c}
\toprule
\multicolumn{3}{c}{\textbf{Object Detection on SUN RGB-D}} \\
Method & Format & mAP@0.5 \\
\hline
% DSS \cite{song2016deep,hou20193d} & - & 42.1 \\
% MRCNN 2D-3D \cite{he2017mask,hou20193d} & 10.5 & 17.3 \\
% F-PointNet \cite{qi2018frustum} & - & 54.0 \\
BoxNet \cite{qi2019deep} & Point & 25.1 \\
3DETR \cite{misra2021end} & Point & 30.1 \\
3DETR-m \cite{misra2021end} & Point & 32.7 \\
VoteNet \cite{qi2019deep}  & Point & 32.9 \\
% HGNet \cite{chen2020hierarchical} & 34.4 & 61.6 \\
\hline
VoteNet \cite{qi2019deep,xie2020pointcontrast}* & Voxel & 31.7 \\
VoteNet+PPKT~\cite{liu2021learning}* & Voxel & 33.9 (\textcolor{blue}{\scriptsize{+2.2}})  \\
VoteNet+PC-FractalDB~\cite{yamada2022point}$\dag$ & Point & 33.9 (\textcolor{blue}{\scriptsize{+2.2}})  \\
VoteNet+PointContrast~\cite{xie2020pointcontrast}* & Voxel & 34.8 (\textcolor{blue}{\scriptsize{+3.1}})  \\
VoteNet+RandomRooms~\cite{rao2021randomrooms}$\dag$ & Point & 35.4 (\textcolor{blue}{\scriptsize{+3.7}}) \\
VoteNet+PC-FractalDB~\cite{yamada2022point}* & Voxel & 35.9 (\textcolor{blue}{\scriptsize{+4.2}})  \\
\textbf{VoteNet+Text4Point}* & Voxel & \textbf{37.0 (\textcolor{blue}{\scriptsize{+5.3}})} \\
\bottomrule
\end{tabular}}
\end{center}
\label{tab:sunrgbd05}
\vskip -0.3in
\end{table}

\begin{table}[htbp]\small
\setlength{\abovecaptionskip}{0cm} 
\setlength{\belowcaptionskip}{0cm} 
\caption{\textbf{Object detection results on ScanNet.} We report the mean average precision with 0.5 IoU threshold. * corresponds to the model adopts SR-UNet~\cite{choy20194d} as the backbone. $\dag$ corresponds to the model adopts PointNet++~\cite{qi2017pointnet++} as the backbone.}
\newcommand{\tabincell}[2]{\begin{tabular}{@{}#1@{}}#2\end{tabular}}
\begin{center}
\setlength{\tabcolsep}{2.5mm}{
\renewcommand\arraystretch{1.1}
\begin{tabular}{l|c|c}
\toprule
\multicolumn{3}{c}{\textbf{Object Detection on ScanNet}} \\
Method & Format & mAP@0.5 \\
\hline
DSS \cite{song2016deep,hou20193d} & Voxel & 6.8 \\
MRCNN 2D-3D \cite{he2017mask,hou20193d} & Pixel & 10.5 \\
F-PointNet \cite{qi2018frustum,hou20193d} & Point+Pixel & 10.8 \\
GSPN \cite{yi2019gspn} & Point & 17.7 \\
3D-SIS \cite{hou20193d} & Voxel+Pixel & 22.5 \\
VoteNet \cite{qi2019deep}  & Point & 33.5 \\
\hline
VoteNet~\cite{qi2019deep,xie2020pointcontrast}* & Voxel & 35.4 \\
VoteNet+RandomRooms~\cite{rao2021randomrooms}$\dag$ & Point & 36.2 (\textcolor{blue}{\scriptsize{+0.8}}) \\
VoteNet+PC-FractalDB~\cite{yamada2022point}* & Voxel & 37.0 (\textcolor{blue}{\scriptsize{+1.6}})\\
VoteNet+PointContrast~\cite{xie2020pointcontrast}* & Voxel & 38.0 (\textcolor{blue}{\scriptsize{+2.6}}) \\
VoteNet+PC-FractalDB~\cite{yamada2022point}$\dag$ & Point & 38.3 (\textcolor{blue}{\scriptsize{+2.9}})  \\
VoteNet+PPKT~\cite{liu2021learning}* & Voxel & 38.9 (\textcolor{blue}{\scriptsize{+3.5}})  \\
\textbf{VoteNet+Text4Point}* & Voxel & \textbf{39.7 (\textcolor{blue}{\scriptsize{+4.3}})} \\
\bottomrule
\end{tabular}}
\end{center}
\label{tab:scannet}
\vskip -0.3in
\end{table}

\subsection{Object Detection}
\noindent\textbf{Setup.}
We conduct experiments on two detection benchmarks, SUN RGB-D~\cite{song2015sun} and ScanNet~\cite{dai2017scannet}. SUN RGB-D contains 5K training images annotated with 3D oriented bounding boxes for 37 object categories. ScanNet has 1513 labeled scenes with 40 semantic classes. We adopt the VoteNet~\cite{qi2019deep} framework and switch its original backbone PointNet++~\cite{qi2017pointnet++} to SR-UNet following PointContrast~\cite{xie2020pointcontrast}. Please refer to the Appendix for more training details.
We validate the performance on the 10 classes of SUN RGB-D and 18 classes of ScanNet, and compare our approach with other competitive pre-training methods.

\noindent\textbf{Results.}
The results for SUN RGB-D dataset are shown in Tab.~\ref{tab:sunrgbd05}. It can be observed that simply switching the backbone from PointNet++ to SR-UNet leads to performance degradation. 
This is probably because the configurations of VoteNet are tailored for PointNet++. 
However, with Text4Point, SR-UNet obtains an improvement of 5.3\% mAP and surpasses VoteNet by 4.1\% mAP. 
In Tab.~\ref{tab:scannet}, we present the detection results on the ScanNet dataset. 
Text4Point significantly improves the vanilla SR-UNet by 4.3\% mAP and outperforms other competitive baselines.

\begin{table}[htbp]\small
\setlength{\abovecaptionskip}{0cm} 
\setlength{\belowcaptionskip}{0cm} 
\caption{\textbf{Ablation studies on S3DIS semantic segmentation task.} TQM denotes the Text Querying Module. * corresponds to the baseline model trained from scratch. $\dag$ corresponds to the full model we use.}
\newcommand{\tabincell}[2]{\begin{tabular}{@{}#1@{}}#2\end{tabular}}
\begin{center}
\setlength{\tabcolsep}{1mm}{
\renewcommand\arraystretch{1.2}
\begin{tabular}{l|cccc|cc}
\toprule
&\multicolumn{4}{c|}{Components} & \multicolumn{2}{c}{Result}  \\
& TQM & Pre-train & Auxiliary & Prompt & mIoU & mAcc \\
\hline
1*& \xmark & \xmark & \xmark  &\xmark  & 66.7 & 74.4\\
2& \cmark & \xmark & \xmark &\xmark  & 67.0 & 74.9 \\
3& \cmark & \cmark & \xmark &\xmark  & 69.3 & 76.4 \\
4 & \cmark & \cmark & \cmark & \xmark  & 70.1 & 76.7\\
$5^{\dag}$ & \cmark & \cmark & \cmark & \cmark  &\textbf{70.4} & \textbf{77.3}\\
\bottomrule
\end{tabular}} \\
\end{center}
\label{tab:ablation_component}
\vskip -0.3in
\end{table}

\begin{table}[htbp]\small
\setlength{\abovecaptionskip}{0cm} 
\setlength{\belowcaptionskip}{0cm} 
\caption{\textbf{Ablation study on language modality.} We replace the meaningful text embeddings with random learnable parameters and remain all the other settings unchanged. The significant performance degradation strongly demonstrates that the language modality is the main source of improvement of our method.}
\newcommand{\tabincell}[2]{\begin{tabular}{@{}#1@{}}#2\end{tabular}}
\begin{center}
\setlength{\tabcolsep}{2mm}{
\renewcommand\arraystretch{1.2}
\begin{tabular}{l|cc}
\toprule
\multicolumn{1}{c|}{\multirow{2}{*}{Model}} & \multicolumn{2}{c}{Result}  \\ 
                       & mIoU & mAcc \\
\hline
random learnable embeddings & 68.6 & 74.9 \\
meaningful text embeddings & \textbf{70.4} & \textbf{77.3}\\  
\bottomrule
\end{tabular}} \\
\end{center}
\label{tab:ablation_language}
\vskip -0.3in
\end{table}

\subsection{Ablations}
To demonstrate the effectiveness of each component of Text4Point, we conduct detailed experiments on S3DIS semantic segmentation task. For experimental efficiency, we adopt the Minkowski Engine (ME) v5, which provides CUDA accelerations for coordinate management functions.
It is faster than ME v4 but brings some performance degradation. 
Therefore, we only use ME v5 for ablation studies and adopt ME v4 for the main results in this paper to achieve better performance.
Though ME v5 slightly degrades the accuracy of the model, it is still reliable to reflect the difference between different design variants.

As summarized in Tab.~\ref{tab:ablation_component}, we first directly add the Text Querying Module to the SR-UNet to conduct the segmentation task without pre-training. It brings little performance improvement (66.7\% v.s. 67.0\%), which means that the 3D model cannot acquire benefit from the unaligned text embeddings. On the contrary, if we first pre-train the model with 2D-3D contrastive loss and obtain the alignment between 3D features and 
language embeddings, the performance can be significantly improved (66.7\% v.s. 69.3\%). It indicates that the pre-training process is critical, and the 3D model can learn better representations under language guidance. Though there exists a domain gap between the pre-training and fine-tuning dataset, applying auxiliary loss can address this problem well and bring an improvement of 0.8\% (69.3\% v.s. 70.1\%).
Finally, by leveraging prompt learning, the performance is further improved.

To verify whether the language modality plays an important role in guiding the 3D feature extraction, we replace the meaningful text embeddings generated by the category names of the corresponding datasets with randomly-initialized learnable parameters and remain all the other settings unchanged for both pre-training and fine-tuning. As presented in Tab.~\ref{tab:ablation_language}, this leads to a significant performance degradation (-1.8\%), which strongly demonstrates that the language modality does provide useful guidance to facilitate the 3D representation learning and is the main source of improvement of our method.

\section{Conclusion}
\label{sec:conclusion}
In this paper, we proposed a novel Text4Point framework to bridge the gap between text and point cloud representations via 2D images. For the first time, we introduced language information to guide the 3D dense prediction tasks. 
To integrate language modality with 3D features, we proposed a Text Querying Module that queries text embeddings with point cloud features. 
Extensive experiments on several dense prediction tasks, including point cloud semantic segmentation, instance segmentation, and object detection, validated the effectiveness of our approach.

%%%%%%%%% REFERENCES
{\small
\bibliographystyle{ieee_fullname}
\bibliography{egbib}

\begin{thebibliography}{10}\itemsep=-1pt

\bibitem{anderson2018bottom}
Peter Anderson, Xiaodong He, Chris Buehler, Damien Teney, Mark Johnson, Stephen
  Gould, and Lei Zhang.
\newblock Bottom-up and top-down attention for image captioning and visual
  question answering.
\newblock In {\em CVPR}, 2018.

\bibitem{antol2015vqa}
Stanislaw Antol, Aishwarya Agrawal, Jiasen Lu, Margaret Mitchell, Dhruv Batra,
  C~Lawrence Zitnick, and Devi Parikh.
\newblock Vqa: Visual question answering.
\newblock In {\em ICCV}, 2015.

\bibitem{armeni20163d}
Iro Armeni, Ozan Sener, Amir~R Zamir, Helen Jiang, Ioannis Brilakis, Martin
  Fischer, and Silvio Savarese.
\newblock {3D} semantic parsing of large-scale indoor spaces.
\newblock In {\em CVPR}, 2016.

\bibitem{atzmon2018point}
Matan Atzmon, Haggai Maron, and Yaron Lipman.
\newblock Point convolutional neural networks by extension operators.
\newblock {\em ACM TOG}, 2018.

\bibitem{chen2022self}
Nenglun Chen, Lei Chu, Hao Pan, Yan Lu, and Wenping Wang.
\newblock Self-supervised image representation learning with geometric set
  consistency.
\newblock In {\em CVPR}, 2022.

\bibitem{chen2020uniter}
Yen-Chun Chen, Linjie Li, Licheng Yu, Ahmed El~Kholy, Faisal Ahmed, Zhe Gan, Yu
  Cheng, and Jingjing Liu.
\newblock Uniter: Universal image-text representation learning.
\newblock In {\em ECCV}, 2020.

\bibitem{chen2022autoalign}
Zehui Chen, Zhenyu Li, Shiquan Zhang, Liangji Fang, Qinghong Jiang, Feng Zhao,
  Bolei Zhou, and Hang Zhao.
\newblock Autoalign: Pixel-instance feature aggregation for multi-modal {3D}
  object detection.
\newblock {\em arXiv preprint arXiv:2201.06493}, 2022.

\bibitem{choy20194d}
Christopher Choy, JunYoung Gwak, and Silvio Savarese.
\newblock {4D} spatio-temporal convnets: Minkowski convolutional neural
  networks.
\newblock In {\em CVPR}, 2019.

\bibitem{dai2017scannet}
Angela Dai, Angel~X Chang, Manolis Savva, Maciej Halber, Thomas Funkhouser, and
  Matthias Nie{\ss}ner.
\newblock Scannet: Richly-annotated {3D} reconstructions of indoor scenes.
\newblock In {\em CVPR}, 2017.

\bibitem{gao2021clip}
Peng Gao, Shijie Geng, Renrui Zhang, Teli Ma, Rongyao Fang, Yongfeng Zhang,
  Hongsheng Li, and Yu Qiao.
\newblock Clip-adapter: Better vision-language models with feature adapters.
\newblock {\em arXiv preprint arXiv:2110.04544}, 2021.

\bibitem{gu2021open}
Xiuye Gu, Tsung-Yi Lin, Weicheng Kuo, and Yin Cui.
\newblock Open-vocabulary object detection via vision and language knowledge
  distillation.
\newblock In {\em ICLR}, 2021.

\bibitem{guo2021pct}
Meng-Hao Guo, Jun-Xiong Cai, Zheng-Ning Liu, Tai-Jiang Mu, Ralph~R Martin, and
  Shi-Min Hu.
\newblock Pct: Point cloud transformer.
\newblock {\em Computational Visual Media}, 2021.

\bibitem{he2017mask}
Kaiming He, Georgia Gkioxari, Piotr Doll{\'a}r, and Ross Girshick.
\newblock Mask r-cnn.
\newblock In {\em ICCV}, 2017.

\bibitem{he2016deep}
Kaiming He, Xiangyu Zhang, Shaoqing Ren, and Jian Sun.
\newblock Deep residual learning for image recognition.
\newblock In {\em CVPR}, 2016.

\bibitem{hou20193d}
Ji Hou, Angela Dai, and Matthias Nie{\ss}ner.
\newblock {3D-SIS}: {3D} semantic instance segmentation of {RGB-D} scans.
\newblock In {\em CVPR}, 2019.

\bibitem{hou2021pri3d}
Ji Hou, Saining Xie, Benjamin Graham, Angela Dai, and Matthias Nie{\ss}ner.
\newblock {Pri3D}: Can {3D} priors help {2D} representation learning?
\newblock In {\em CVPR}, 2021.

\bibitem{hu2019revisiting}
Junjie Hu, Mete Ozay, Yan Zhang, and Takayuki Okatani.
\newblock Revisiting single image depth estimation: Toward higher resolution
  maps with accurate object boundaries.
\newblock In {\em WACV}, 2019.

\bibitem{ioffe2015batch}
Sergey Ioffe and Christian Szegedy.
\newblock Batch normalization: Accelerating deep network training by reducing
  internal covariate shift.
\newblock In {\em ICML}, 2015.

\bibitem{jia2021scaling}
Chao Jia, Yinfei Yang, Ye Xia, Yi-Ting Chen, Zarana Parekh, Hieu Pham, Quoc Le,
  Yun-Hsuan Sung, Zhen Li, and Tom Duerig.
\newblock Scaling up visual and vision-language representation learning with
  noisy text supervision.
\newblock In {\em ICML}, 2021.

\bibitem{jiang2020pointgroup}
Li Jiang, Hengshuang Zhao, Shaoshuai Shi, Shu Liu, Chi-Wing Fu, and Jiaya Jia.
\newblock Pointgroup: Dual-set point grouping for {3D} instance segmentation.
\newblock In {\em CVPR}, 2020.

\bibitem{kwon2022clipstyler}
Gihyun Kwon and Jong~Chul Ye.
\newblock Clipstyler: Image style transfer with a single text condition.
\newblock In {\em CVPR}, 2022.

\bibitem{landrieu2018large}
Loic Landrieu and Martin Simonovsky.
\newblock Large-scale point cloud semantic segmentation with superpoint graphs.
\newblock In {\em CVPR}, 2018.

\bibitem{li2018pointcnn}
Yangyan Li, Rui Bu, Mingchao Sun, Wei Wu, Xinhan Di, and Baoquan Chen.
\newblock Pointcnn: Convolution on x-transformed points.
\newblock In {\em NeurIPS}, 2018.

\bibitem{liang2021instance}
Zhihao Liang, Zhihao Li, Songcen Xu, Mingkui Tan, and Kui Jia.
\newblock Instance segmentation in {3D} scenes using semantic superpoint tree
  networks.
\newblock In {\em ICCV}, 2021.

\bibitem{liu2022open}
Quande Liu, Youpeng Wen, Jianhua Han, Chunjing Xu, Hang Xu, and Xiaodan Liang.
\newblock Open-world semantic segmentation via contrasting and clustering
  vision-language embedding.
\newblock In {\em ECCV}, 2022.

\bibitem{liu2021learning}
Yueh-Cheng Liu, Yu-Kai Huang, Hung-Yueh Chiang, Hung-Ting Su, Zhe-Yu Liu,
  Chin-Tang Chen, Ching-Yu Tseng, and Winston~H Hsu.
\newblock Learning from {2D}: Contrastive pixel-to-point knowledge transfer for
  {3D} pretraining.
\newblock {\em arXiv preprint arXiv:2104.04687}, 2021.

\bibitem{maturana2015voxnet}
Daniel Maturana and Sebastian Scherer.
\newblock Voxnet: A {3D} convolutional neural network for real-time object
  recognition.
\newblock In {\em IROS}, 2015.

\bibitem{misra2021end}
Ishan Misra, Rohit Girdhar, and Armand Joulin.
\newblock An end-to-end transformer model for {3D} object detection.
\newblock In {\em ICCV}, 2021.

\bibitem{mokady2021clipcap}
Ron Mokady, Amir Hertz, and Amit~H Bermano.
\newblock Clipcap: Clip prefix for image captioning.
\newblock {\em arXiv preprint arXiv:2111.09734}, 2021.

\bibitem{patashnik2021styleclip}
Or Patashnik, Zongze Wu, Eli Shechtman, Daniel Cohen-Or, and Dani Lischinski.
\newblock Styleclip: Text-driven manipulation of stylegan imagery.
\newblock In {\em ICCV}, 2021.

\bibitem{qi2020imvotenet}
Charles~R Qi, Xinlei Chen, Or Litany, and Leonidas~J Guibas.
\newblock Imvotenet: Boosting {3D} object detection in point clouds with image
  votes.
\newblock In {\em CVPR}, 2020.

\bibitem{qi2019deep}
Charles~R Qi, Or Litany, Kaiming He, and Leonidas~J Guibas.
\newblock Deep hough voting for {3D} object detection in point clouds.
\newblock In {\em ICCV}, 2019.

\bibitem{qi2018frustum}
Charles~R Qi, Wei Liu, Chenxia Wu, Hao Su, and Leonidas~J Guibas.
\newblock Frustum pointnets for {3D} object detection from {RGB-D} data.
\newblock In {\em CVPR}, 2018.

\bibitem{qi2017pointnet}
Charles~R Qi, Hao Su, Kaichun Mo, and Leonidas~J Guibas.
\newblock Pointnet: Deep learning on point sets for {3D} classification and
  segmentation.
\newblock In {\em CVPR}, 2017.

\bibitem{qi2017pointnet++}
Charles~Ruizhongtai Qi, Li Yi, Hao Su, and Leonidas~J Guibas.
\newblock Pointnet++: Deep hierarchical feature learning on point sets in a
  metric space.
\newblock In {\em NeurIPS}, 2017.

\bibitem{radford2021learning}
Alec Radford, Jong~Wook Kim, Chris Hallacy, Aditya Ramesh, Gabriel Goh,
  Sandhini Agarwal, Girish Sastry, Amanda Askell, Pamela Mishkin, Jack Clark,
  et~al.
\newblock Learning transferable visual models from natural language
  supervision.
\newblock In {\em ICML}, 2021.

\bibitem{rao2021randomrooms}
Yongming Rao, Benlin Liu, Yi Wei, Jiwen Lu, Cho-Jui Hsieh, and Jie Zhou.
\newblock Randomrooms: Unsupervised pre-training from synthetic shapes and
  randomized layouts for {3D} object detection.
\newblock In {\em ICCV}, 2021.

\bibitem{rao2021denseclip}
Yongming Rao, Wenliang Zhao, Guangyi Chen, Yansong Tang, Zheng Zhu, Guan Huang,
  Jie Zhou, and Jiwen Lu.
\newblock Denseclip: Language-guided dense prediction with context-aware
  prompting.
\newblock In {\em CVPR}, 2022.

\bibitem{ronneberger2015unet}
Olaf Ronneberger, Philipp Fischer, and Thomas Brox.
\newblock U-net: Convolutional networks for biomedical image segmentation.
\newblock In {\em MICCAI}, 2015.

\bibitem{song2015sun}
Shuran Song, Samuel~P Lichtenberg, and Jianxiong Xiao.
\newblock {SUN RGB-D}: A {RGB-D} scene understanding benchmark suite.
\newblock In {\em CVPR}, 2015.

\bibitem{song2016deep}
Shuran Song and Jianxiong Xiao.
\newblock Deep sliding shapes for amodal {3D} object detection in {RGB-D}
  images.
\newblock In {\em CVPR}, 2016.

\bibitem{su2019vl}
Weijie Su, Xizhou Zhu, Yue Cao, Bin Li, Lewei Lu, Furu Wei, and Jifeng Dai.
\newblock Vl-bert: Pre-training of generic visual-linguistic representations.
\newblock In {\em ICLR}, 2019.

\bibitem{tatarchenko2018tangent}
Maxim Tatarchenko, Jaesik Park, Vladlen Koltun, and Qian-Yi Zhou.
\newblock Tangent convolutions for dense prediction in {3D}.
\newblock In {\em CVPR}, 2018.

\bibitem{tchapmi2017segcloud}
Lyne Tchapmi, Christopher Choy, Iro Armeni, JunYoung Gwak, and Silvio Savarese.
\newblock Segcloud: Semantic segmentation of {3D} point clouds.
\newblock In {\em 3DV}, 2017.

\bibitem{thomas2019kpconv}
Hugues Thomas, Charles~R Qi, Jean-Emmanuel Deschaud, Beatriz Marcotegui,
  Fran{\c{c}}ois Goulette, and Leonidas~J Guibas.
\newblock Kpconv: Flexible and deformable convolution for point clouds.
\newblock In {\em CVPR}, 2019.

\bibitem{uy2019revisiting}
Mikaela~Angelina Uy, Quang-Hieu Pham, Binh-Son Hua, Thanh Nguyen, and Sai-Kit
  Yeung.
\newblock Revisiting point cloud classification: A new benchmark dataset and
  classification model on real-world data.
\newblock In {\em CVPR}, 2019.

\bibitem{wang2021unsupervised}
Hanchen Wang, Qi Liu, Xiangyu Yue, Joan Lasenby, and Matt~J Kusner.
\newblock Unsupervised point cloud pre-training via occlusion completion.
\newblock In {\em ICCV}, 2021.

\bibitem{wang2018sgpn}
Weiyue Wang, Ronald Yu, Qiangui Huang, and Ulrich Neumann.
\newblock Sgpn: Similarity group proposal network for {3D} point cloud instance
  segmentation.
\newblock In {\em CVPR}, 2018.

\bibitem{wang2019associatively}
Xinlong Wang, Shu Liu, Xiaoyong Shen, Chunhua Shen, and Jiaya Jia.
\newblock Associatively segmenting instances and semantics in point clouds.
\newblock In {\em CVPR}, 2019.

\bibitem{wang2019dynamic}
Yue Wang, Yongbin Sun, Ziwei Liu, Sanjay~E Sarma, Michael~M Bronstein, and
  Justin~M Solomon.
\newblock Dynamic graph cnn for learning on point clouds.
\newblock {\em ACM TOG}, 2019.

\bibitem{wang2019camp}
Zihao Wang, Xihui Liu, Hongsheng Li, Lu Sheng, Junjie Yan, Xiaogang Wang, and
  Jing Shao.
\newblock Camp: Cross-modal adaptive message passing for text-image retrieval.
\newblock In {\em CVPR}, 2019.

\bibitem{wu20153d}
Zhirong Wu, Shuran Song, Aditya Khosla, Fisher Yu, Linguang Zhang, Xiaoou Tang,
  and Jianxiong Xiao.
\newblock {3D} shapenets: A deep representation for volumetric shapes.
\newblock In {\em CVPR}, 2015.

\bibitem{xie2020pointcontrast}
Saining Xie, Jiatao Gu, Demi Guo, Charles~R Qi, Leonidas Guibas, and Or Litany.
\newblock Pointcontrast: Unsupervised pre-training for {3D} point cloud
  understanding.
\newblock In {\em ECCV}, 2020.

\bibitem{xu2015show}
Kelvin Xu, Jimmy Ba, Ryan Kiros, Kyunghyun Cho, Aaron Courville, Ruslan
  Salakhudinov, Rich Zemel, and Yoshua Bengio.
\newblock Show, attend and tell: Neural image caption generation with visual
  attention.
\newblock In {\em ICML}, 2015.

\bibitem{yamada2022point}
Ryosuke Yamada, Hirokatsu Kataoka, Naoya Chiba, Yukiyasu Domae, and Tetsuya
  Ogata.
\newblock Point cloud pre-training with natural {3D} structures.
\newblock In {\em CVPR}, 2022.

\bibitem{yan2018second}
Yan Yan, Yuxing Mao, and Bo Li.
\newblock Second: Sparsely embedded convolutional detection.
\newblock {\em Sensors}, 2018.

\bibitem{yang2019learning}
Bo Yang, Jianan Wang, Ronald Clark, Qingyong Hu, Sen Wang, Andrew Markham, and
  Niki Trigoni.
\newblock Learning object bounding boxes for {3D} instance segmentation on
  point clouds.
\newblock In {\em NeurIPS}, 2019.

\bibitem{yang20203dssd}
Zetong Yang, Yanan Sun, Shu Liu, and Jiaya Jia.
\newblock 3dssd: Point-based {3D} single stage object detector.
\newblock In {\em CVPR}, 2020.

\bibitem{ye20183d}
Xiaoqing Ye, Jiamao Li, Hexiao Huang, Liang Du, and Xiaolin Zhang.
\newblock {3D} recurrent neural networks with context fusion for point cloud
  semantic segmentation.
\newblock In {\em ECCV}, 2018.

\bibitem{yi2019gspn}
Li Yi, Wang Zhao, He Wang, Minhyuk Sung, and Leonidas~J Guibas.
\newblock Gspn: Generative shape proposal network for {3D} instance
  segmentation in point cloud.
\newblock In {\em CVPR}, 2019.

\bibitem{yin2021center}
Tianwei Yin, Xingyi Zhou, and Philipp Krahenbuhl.
\newblock Center-based {3D} object detection and tracking.
\newblock In {\em CVPR}, 2021.

\bibitem{yu2022point}
Xumin Yu, Lulu Tang, Yongming Rao, Tiejun Huang, Jie Zhou, and Jiwen Lu.
\newblock Point-bert: Pre-training {3D} point cloud transformers with masked
  point modeling.
\newblock In {\em CVPR}, 2022.

\bibitem{yuan2021florence}
Lu Yuan, Dongdong Chen, Yi-Ling Chen, Noel Codella, Xiyang Dai, Jianfeng Gao,
  Houdong Hu, Xuedong Huang, Boxin Li, Chunyuan Li, et~al.
\newblock Florence: A new foundation model for computer vision.
\newblock {\em arXiv preprint arXiv:2111.11432}, 2021.

\bibitem{zhang2021vinvl}
Pengchuan Zhang, Xiujun Li, Xiaowei Hu, Jianwei Yang, Lei Zhang, Lijuan Wang,
  Yejin Choi, and Jianfeng Gao.
\newblock Vinvl: Revisiting visual representations in vision-language models.
\newblock In {\em CVPR}, 2021.

\bibitem{zhang2021pointclip}
Renrui Zhang, Ziyu Guo, Wei Zhang, Kunchang Li, Xupeng Miao, Bin Cui, Yu Qiao,
  Peng Gao, and Hongsheng Li.
\newblock Pointclip: Point cloud understanding by clip.
\newblock In {\em CVPR}, 2022.

\bibitem{zhao2021point}
Hengshuang Zhao, Li Jiang, Jiaya Jia, Philip~HS Torr, and Vladlen Koltun.
\newblock Point transformer.
\newblock In {\em ICCV}, 2021.

\bibitem{zhong2022regionclip}
Yiwu Zhong, Jianwei Yang, Pengchuan Zhang, Chunyuan Li, Noel Codella,
  Liunian~Harold Li, Luowei Zhou, Xiyang Dai, Lu Yuan, Yin Li, et~al.
\newblock Regionclip: Region-based language-image pretraining.
\newblock In {\em CVPR}, 2022.

\bibitem{zhou2022extract}
Chong Zhou, Chen~Change Loy, and Bo Dai.
\newblock Extract free dense labels from clip.
\newblock In {\em ECCV}, 2022.

\bibitem{zhou2021learning}
Kaiyang Zhou, Jingkang Yang, Chen~Change Loy, and Ziwei Liu.
\newblock Learning to prompt for vision-language models.
\newblock {\em IJCV}, 2022.

\bibitem{zhou2018voxelnet}
Yin Zhou and Oncel Tuzel.
\newblock Voxelnet: End-to-end learning for point cloud based {3D} object
  detection.
\newblock In {\em CVPR}, 2018.

\bibitem{zhou2022towards}
Yufan Zhou, Ruiyi Zhang, Changyou Chen, Chunyuan Li, Chris Tensmeyer, Tong Yu,
  Jiuxiang Gu, Jinhui Xu, and Tong Sun.
\newblock Towards language-free training for text-to-image generation.
\newblock In {\em CVPR}, 2022.

\end{thebibliography}
}
\clearpage
\newpage
\appendix
\section*{Appendix}
\section{Detailed Results of Object Detection}
We provide per-category average precision performance with 0.5 IoU threshold on SUN RGB-D~\cite{song2015sun} and ScanNet datasets~\cite{dai2017scannet} in Tab.~\ref{tab:sunrgbd05_detail} and Tab.~\ref{tab:scannet05_detail}, respectively. On SUN RGB-D, Text4Point significantly improves the vanilla SR-UNet~\cite{choy20194d} by 5.3\% mAP and surpasses it in 8 out of 10 categories. On ScanNet, Text4Point brings an improvement of 4.3\% mAP for the vanilla SR-UNet and outperforms it in 12 out of 18 categories.

\begin{table*}[hb]\small
\setlength{\abovecaptionskip}{0cm} 
\setlength{\belowcaptionskip}{-0.2cm} 
\caption{\textbf{Object detection results on SUN RGB-D dataset.} We report per-category  average precision performance with 0.5 IoU threshold. * corresponds to the model adopts SR-UNet~\cite{choy20194d} as the backbone.}
% \vskip -0.05in
\newcommand{\tabincell}[2]{\begin{tabular}{@{}#1@{}}#2\end{tabular}}
\begin{center}
\renewcommand\arraystretch{1.1}
\begin{tabular}{l|cccccccccc|c}
\toprule
\multicolumn{12}{c}{\textbf{Object Detection on SUN RGB-D}} \\
Method & bed& table& sofa& chair &toilet &desk& dresser &nightstd& bkshlf& bathtub & mAP \\
\hline
VoteNet \cite{qi2019deep} & 47.3 & 19.7  & 41.4  &  54.1 & 58.6  & 5.2  &  13.6  &  35.0 & 4.6  & 49.9   & 32.9  \\
\hline
VoteNet* \cite{qi2019deep,choy20194d} & 47.8 & 19.6 & 48.1 & 54.6 & 60.0&  6.3 & 15.8&  27.3&  5.4 & 32.1  &31.7\\
VoteNet+PPKT \cite{liu2021learning}* &52.1& 17.6& 50.3 &52.3& 58.1& 6.2 &13.0& 43.7 &9.4& 36.6& 33.9\\
VoteNet+PointContrast \cite{xie2020pointcontrast}* & 50.5& 19.4 &51.8 &54.9& 57.4& 7.5& 16.2& 37.0 &5.9& 47.6 &34.8   \\
\textbf{VoteNet+Text4Point}* & 52.7 &  20.2 & 52.5  & 52.2  &  64.1 &  6.6 &  15.4  &  43.0  &  8.4  & 54.7  &  \textbf{37.0}\\
\bottomrule
\end{tabular}
% }
\end{center}
\label{tab:sunrgbd05_detail}
% \vskip -0.25in
\end{table*}

\begin{table*}[hb]\small
\setlength{\abovecaptionskip}{0cm} 
\setlength{\belowcaptionskip}{-0.5cm} 
\caption{\textbf{Object detection results on ScanNet dataset.}  We report per-category  average precision performance with 0.5 IoU threshold. * corresponds to the model adopts SR-UNet~\cite{choy20194d} as the backbone.}
% \vskip -0.05in
\newcommand{\tabincell}[2]{\begin{tabular}{@{}#1@{}}#2\end{tabular}}
\begin{center}
\setlength{\tabcolsep}{0.4mm}{
\renewcommand\arraystretch{1.2}
\begin{tabular}{l|cccccccccccccccccc|c}
\toprule
\multicolumn{20}{c}{\textbf{Object Detection on ScanNet}} \\
Method &   cab &   bed&   chair &   sofa &   table&   door &   wind &   bkshlf &   pic&   cntr &   desk &   curtain&   refrig &   shower &   toilet&   sink &   bath &   garbage & mAP \\
\hline
VoteNet \cite{qi2019deep}& 8.1& 76.1 &67.2& 68.8& 42.4& 15.3& 6.4& 28.0 &1.3 &9.5 &37.5& 11.6& 27.8& 10.0 &86.5& 16.8 &78.9& 11.7 &33.5\\
\hline
VoteNet* \cite{qi2019deep,choy20194d}& 9.9 &70.5& 70.0 &60.5 &43.4 &21.8& 10.5 &33.3& 0.8& 15.4 &33.3& 26.6& 39.3 &9.7 &74.7 &23.7 &75.8 &18.1 &35.4\\
VoteNet+PointContrast \cite{xie2020pointcontrast}*& 13.1 &74.7& 75.4 &61.3 &44.8& 19.8& 12.9 &32.0&0.9 &21.9& 31.9& 27.0 &32.6& 17.5& 87.4& 23.2& 80.8 &26.7 &38.0\\
VoteNet+PPKT \cite{liu2021learning}* &8.0&  70.3 & 72.6&  57.0&  40.3&  23.3&  11.4&  45.3 & 3.5 & 18.8 & 38.1 & 25.1&  31.3&  35.5&  86.7&  18.1&  87.1 & 28.1&  38.9\\
\textbf{VoteNet+Text4Point}* & 8.0 & 72.6  & 69.5  & 61.3  & 44.1  & 21.7 & 14.6 &  38.2  & 1.9 &  13.4  &  35.9 & 23.7  &  31.7  &  53.1 &  90.2  & 26.3  & 79.2& 29.3  &  \textbf{39.7} \\
\bottomrule
\end{tabular}}
\end{center}
\label{tab:scannet05_detail}
% \vskip -0.25in
\end{table*}

\section{Implementation Details}
\subsection{Semantic and Instance Segmentation}
Following \cite{xie2020pointcontrast}, we use the data augmentation techniques such as random horizontal flip, color jittering, random rotation, as well as scale augmentation. The voxel size is set to 5cm.
We use a momentum SGD optimizer with an initial learning rate of 0.1 and adopt a polynomial scheduler with a power factor of 0.9. Weight decay is set to 0.0001. The model is trained for 60k iterations with a batch size of 48. We set the temperature for auxiliary loss to 0.07. The length of learnable contexts for prompt learning is 8.

\subsection{Object Detection}
Following \cite{xie2020pointcontrast}, we use VoteNet~\cite{qi2019deep} framework by switching its original backbone PointNet++~\cite{qi2017pointnet++} to SR-UNet and adopting most of the configurations.
We sample 20,000 points for the SUN RGB-D dataset~\cite{song2015sun} and 40,000 points for the ScanNet dataset~\cite{dai2017scannet}. The voxel size is 2.5cm for both SUN RGB-D and ScanNet datasets. 
For SUN RGB-D, we train the model with a batch size of 64 for 180 epochs. 
For ScanNet, the model is trained with a batch size of 32 and learning rate warmup of 15 epochs. 
The initial learning rate is set to 0.001 and decayed by a cosine scheduler.
The temperature for the auxiliary loss is 0.07. The length of learnable contexts for prompt learning is 8.

\section{Ablation Study on the Depth Loss}
To provide meaningful 2D features for the 3D decoder to align, we supervise the 2D decoder by depth prediction~\cite{hu2019revisiting} using the free depth images in the RGB-D dataset during the pre-training stage.
We investigate the impact of utilizing the auxiliary depth loss on the S3DIS~\cite{armeni20163d} semantic segmentation task.
For experimental efficiency, we use Minkowski Engine (ME) v5~\cite{choy20194d}, which provides CUDA accelerations. Though ME v5 slightly degrades the accuracy of the model compared with ME v4, it is still reliable to reflect the difference between different settings. With the depth loss, the 3D model can be better pre-trained, which boosts the performance on the downstream task (+0.7\% mIoU). Thus, we adopt it during pre-training by default.

\end{document}